\documentclass[runningheads]{llncs}

\usepackage{eccv}

\usepackage{eccvabbrv}

\usepackage{graphicx}
\usepackage{booktabs}
\usepackage{microtype}      
\usepackage{xcolor}         
\usepackage{array}
\usepackage{adjustbox}
\usepackage{multirow}
\usepackage{hhline}
\usepackage{makecell}
\usepackage{textcomp}
\usepackage{booktabs}
\usepackage{amsmath}
\usepackage{amssymb}
\usepackage{graphicx}
\usepackage{subcaption}
\usepackage{afterpage}
\usepackage{placeins}
\usepackage{colortbl}
\usepackage{tabularx}
\usepackage{xcolor}
\usepackage{float}

\usepackage{siunitx}
\usepackage[table]{xcolor}

\usepackage{caption}
\usepackage[accsupp]{axessibility}  

\usepackage{hyperref}

\begin{document}

\title{Horizon3D: Sparse Radar-Camera Fusion for Long-Range 3D Perception in Autonomous Driving} 

\titlerunning{Horizon3D}
\author{Geonho Bang~\and
Geunju Baek~\and
Dongyoung Lee~\and
Wonjun Jeong~\and
\\Jun Won Choi\thanks{Corresponding author.}
}

\authorrunning{G. Bang et al.}

\institute{
Seoul National University, Seoul, Republic of Korea\\
\email{\{ghbang,gjbaek,dylee,wjjeong\}@adr.snu.ac.kr, junwchoi@snu.ac.kr}
\url{https://geonhobang.github.io/horizon3d-project-page/}
}

\maketitle

\begin{abstract}
  Long-range 3D object detection is critical for safe autonomous driving at highway speeds, yet existing radar-camera fusion methods face notable limitations at extended ranges. 
  BEV-based approaches effectively encode scene-level context but incur rapidly growing computational cost and struggle to preserve fine-grained object-level detail, while query-based methods provide efficient object-centric encoding but lack sufficient scene-level context. 
  Temporal fusion introduces additional challenges: distant objects produce only a few radar returns and occupy only a few image pixels, requiring scene-level accumulation, while high-speed motion causes large inter-frame displacements that require object-level motion modeling. 
  BEV-based aggregation alleviates sparsity through multi-frame accumulation but is less suited to individual object motion, whereas query-based modeling captures object-level motion but provides limited scene-level temporal context. 
  In this paper, we propose Horizon3D, a sparse radar-camera fusion framework for long-range 3D object detection that jointly captures object-level detail and scene-level context in both spatial and temporal dimensions through a hybrid representation that combines Gaussian primitives with sparse BEV features.
  Horizon3D first employs Keypoint-Guided Gaussian Initialization (KGGI) to initialize Gaussian primitives at object keypoints estimated from radar and camera features. 
  Object-Centric Sparse Fusion (OCSF) aggregates cross-modal features around these primitives and splats the refined Gaussians onto the BEV plane, where they are fused with sparse radar BEV features to combine object-level detail with scene-level context. 
  Finally, Dual-Path Temporal Fusion (DPTF) aggregates temporal cues through a BEV path for multi-frame feature accumulation and a Gaussian path for propagating primitives across frames to encode per-object motion. 
  Extensive evaluations on TruckScenes demonstrate that Horizon3D achieves state-of-the-art performance for radar-camera 3D object detection. On the validation set, our approach outperforms the previous best method by $\mathbf{+3.0}$ NDS and $\mathbf{+1.6}$ mAP while maintaining a sparse representation with competitive inference speed.
\end{abstract}

\begin{center}
    \centering
    \includegraphics[width=0.99\textwidth]{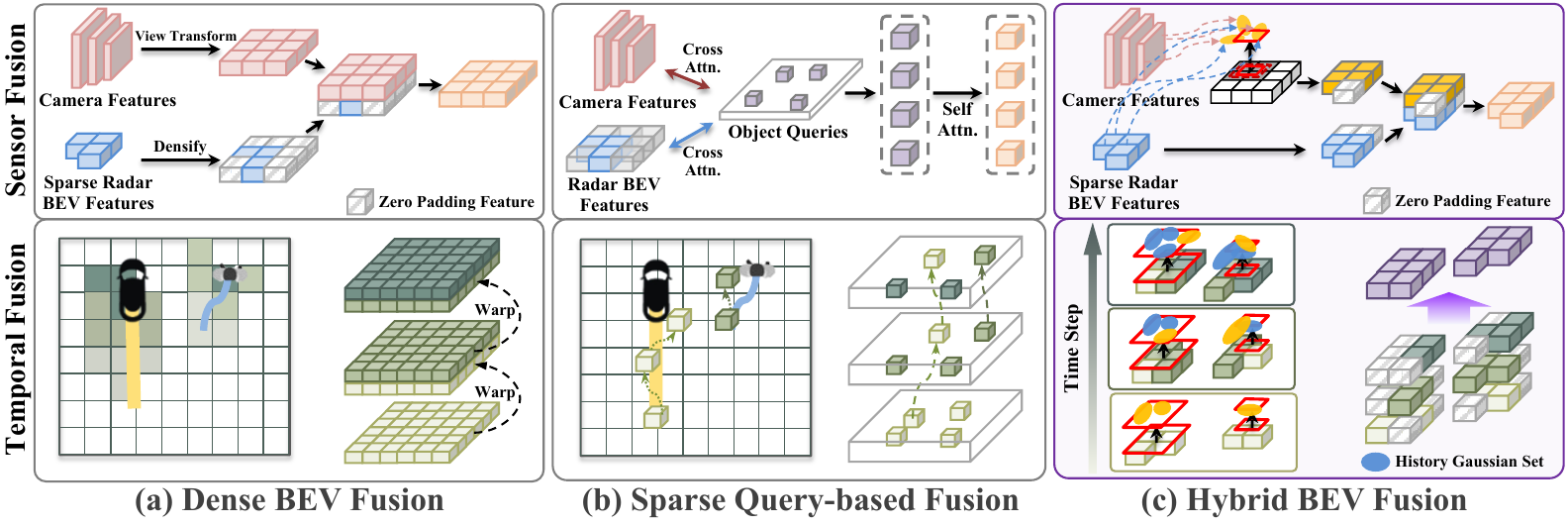}
    \captionof{figure}{\textbf{Comparison of radar-camera fusion paradigms.} (a) Dense BEV fusion encodes scene-level context but incurs growing cost with range, and its uniform temporal aggregation struggles to capture per-object motion.
    (b) Sparse query-based fusion achieves efficient object-centric encoding but lacks scene-level context in the spatiotemporal domain.
    (c) Our hybrid BEV fusion employs Gaussian primitives and sparse BEV features for object- and scene-level encoding, and uses dual-path temporal fusion.}
    \label{fig:Fig1}
\end{center}

\section{Introduction}
\label{sec:intro}

Reliable long-range perception is essential for safe autonomous driving, as substantially longer braking distances at highway speeds require accurate detection of objects beyond $150\,\mathrm{m}$ \cite{truckscenes,ZOD,LR3D,AV2}.
Prior work on long-range 3D object detection has predominantly focused on LiDAR-based methods \cite{FSHNet, DSVT, VoxelMamba}. However, the high deployment and maintenance costs of LiDAR, along with its vulnerability to adverse weather, limit its practicality for large-scale deployment.
As a cost-effective alternative, radar-camera fusion methods \cite{RCBEVDet, CRN, CRT-Fusion, RCM-Fusion, RCTrans, rctdistill} have gained significant attention. Recent work has increasingly adopted 4D radar, which provides spatial measurements including elevation, remains robust to adverse weather, and offers velocity cues for dynamic objects \cite{CVFusion,LXL,SGDet3D}.
Despite this progress, widely used benchmarks (e.g., nuScenes, VoD, and TJ4D) \cite{nuscenes,VoD,TJ4D} primarily target short- to mid-range urban perception.
Although TruckScenes \cite{truckscenes} was introduced for long-range, high-speed scenarios, fully exploiting 4D radar-camera complementarity in this setting remains largely unexplored.

Existing radar-camera 3D object detection frameworks can be broadly categorized into BEV-based and query-based methods. 
As illustrated in Fig.~\ref{fig:Fig1}, both paradigms exhibit clear limitations in long-range detection, especially in sensor fusion and temporal fusion.
For sensor fusion, BEV-based methods~\cite{CRN,RCBEVDet,RCM-Fusion,CRT-Fusion,LXL,SGDet3D,CVFusion} project multimodal features into a unified dense BEV representation, which effectively captures scene-level context but incurs rapidly increasing computational cost as the detection range grows. Moreover, the dense BEV grid encodes the scene uniformly at a fixed spatial resolution, limiting fine-grained object-level detail (Fig.~\ref{fig:Fig1}(a)). 
Query-based methods~\cite{RaCFormer,RCTrans,SpaRC} instead perform object-centric encoding around learnable queries, enabling efficient long-range perception. 
However, the queries are optimized to represent foreground objects, and the resulting representation lacks sufficient scene-level context \cite{bevnext, MambaFusion} (Fig.~\ref{fig:Fig1}(b)).

In this paper, we tackle the challenges of modeling both scene-level and object-level information for temporal fusion in long-range 3D perception.
First, long-range objects produce sparse radar returns and occupy only a few image pixels, making single-frame modeling insufficient for robust perception. Consequently, an effective temporal fusion strategy is required to aggregate complementary information across multiple frames.
Second, dynamic objects exhibit substantial motion between consecutive frames, particularly at highway speeds. Therefore, temporal feature propagation should be performed at the object level to accurately capture the motion of individual objects over time.

Several temporal fusion methods have been proposed for 3D perception. 
BEV-based methods~\cite{CRN,RCBEVDet} aggregate dense BEV features over the entire spatial grid, effectively mitigating the sparsity of single-frame observations through multi-frame fusion. Although methods such as CRT-Fusion~\cite{CRT-Fusion} introduce local motion compensation, they still rely on dense BEV representations, resulting in high computational cost, particularly for long-range perception.
In contrast, query-based methods~\cite{RCTrans,SpaRC} propagate compact object queries across frames, enabling each query to capture the temporal evolution of an individual object. While this design efficiently models object motion, it provides limited scene-level temporal context.
Therefore, effective long-range 3D perception requires a temporal fusion method that jointly models scene-level context and object-level dynamics while maintaining high computational efficiency for extended-range perception.

In this paper, we propose Horizon3D, a sparse radar-camera fusion framework for long-range 3D object detection that addresses these challenges through a hybrid representation combining Gaussian primitives and sparse BEV features (Fig.~\ref{fig:Fig1}(c)).
Our approach comprises three key modules. 
The Keypoint-Guided Gaussian Initialization (KGGI) module estimates object keypoints from both radar and camera features and initializes a compact set of Gaussian primitives at these locations to guide object-centric encoding.
Unlike existing Gaussian-based methods~\cite{GaussianFormer, GaussianFormerV2} that locate numerous primitives across the scene, KGGI places primitives only at estimated object locations, enabling efficient computation even at extended detection ranges.
Given the initialized Gaussians, the Object-Centric Sparse Fusion (OCSF) module aggregates cross-modal features around these primitives and iteratively refines their parameters to align with object structures. 
The refined Gaussians are then splatted onto the BEV plane to produce a sparse object-centric BEV representation, which is fused with sparse radar BEV features to encode fine-grained object-level detail and scene-level context.
The Dual-Path Temporal Fusion (DPTF) module integrates temporal information through two complementary paths.
The BEV path accumulates foreground features over multiple frames to compensate for the per-frame sparsity of distant objects, while the Gaussian path propagates past Gaussians to the current frame and  encodes per-object motion across frames.
In both paths, predicted velocities are used to compensate for the motion of dynamic objects, mitigating temporal misalignment in high-speed scenarios.

We evaluate Horizon3D on the TruckScenes benchmark~\cite{truckscenes}, which features high-speed and long-range driving scenarios. On the validation split, Horizon3D outperforms the previous best radar-camera fusion method by $\mathbf{+3.0}$ NDS and $\mathbf{+1.6}$ mAP, achieving state-of-the-art performance while running faster than existing BEV-based fusion methods.

The main contributions of this paper are as follows:
\begin{itemize}
    \item We propose \textbf{Horizon3D}, a sparse radar-camera fusion framework for long-range 3D object detection that combines Gaussian primitives and sparse BEV features into an efficient hybrid representation.
    
    \item We introduce \textbf{KGGI} and \textbf{OCSF}. KGGI initializes Gaussian primitives at estimated object keypoints, while OCSF iteratively refines them to construct a sparse BEV representation that jointly captures object-level structure and scene-level context without requiring dense BEV construction. As a result, our approach significantly reduces computational cost.
    
    \item We propose \textbf{DPTF}, a dual-path temporal fusion that simultaneously addresses per-frame sparsity at long range and object motion at high speed. It employs  velocity-based compensation in both paths to maintain temporal consistency.
    
    \item Extensive experiments on \textbf{TruckScenes} demonstrate that Horizon3D achieves state-of-the-art performance among radar-camera fusion methods, outperforming the previous best method by $\mathbf{+3.0}$ NDS and $\mathbf{+1.6}$ mAP while maintaining a sparse representation with competitive inference speed.

\end{itemize}

\section{Related Work}

\subsection{Radar-Camera 3D Object Detection}
Radar-camera fusion for 3D object detection combines camera semantics with radar's robust range and velocity measurements. Recent work has increasingly adopted 4D radar, which additionally provides elevation information \cite{CVFusion, LXL, SGDet3D}. Prior methods can be broadly categorized into BEV-based and query-based approaches.

BEV-based methods project multimodal features into a unified BEV space to align cross-sensor features and capture scene-level context \cite{CRN, RCM-Fusion, RCBEVDet, CRT-Fusion}. Radar measurements can serve as depth priors for view transformation \cite{CRN, LXL}, and cross-attention or multi-level refinement can be applied for flexible feature alignment \cite{RCBEVDet, RCM-Fusion}. In contrast, query-based methods aggregate multimodal features using a limited number of object queries, constraining computational growth at longer ranges \cite{RCTrans, RaCFormer, SpaRC}. Recent methods use range-aware circular query initialization \cite{RaCFormer}, densify sparse radar features for query-centric fusion \cite{RCTrans}, or perform alignment directly in a sparse feature space \cite{SpaRC}. However, BEV-based methods incur rapidly growing computational cost as the detection range extends, while query-based methods often lack sufficient scene-level context.

Temporal fusion has been widely adopted to improve detection by aggregating consecutive frames \cite{CRT-Fusion, RCBEVDet, RCTrans, SpaRC}, and can further compensate for per-frame sparsity and large inter-frame displacements at extended ranges. 
BEV-based methods accumulate ego-motion-compensated BEV features over multiple frames \cite{CRT-Fusion,RCBEVDet}.
CRT-Fusion \cite{CRT-Fusion} further addresses the misalignment of moving objects by explicitly modeling object-level dynamics, but still relies on dense BEV aggregation.
Query-based methods propagate object queries across frames to aggregate temporal information \cite{RCTrans, SpaRC}, yet their object-centric aggregation provides limited scene-level temporal context. 
However, temporal fusion that efficiently addresses both per-frame sparsity and object-level motion at long range without dense BEV construction remains unexplored.

\subsection{Long-range 3D Object Detection}
Long-range 3D object detection faces compounded challenges as the sensing range extends, since observations degrade while representation costs grow. Recent progress in long-range detection has been driven largely by LiDAR-based methods, which adopt a fully sparse paradigm that focuses computation on non-empty regions \cite{FSD,voxelnext,SAFDNet}.
Sparse convolutions selectively process occupied voxels to mitigate computational growth \cite{FSD,voxelnext}.
To compensate for the limited context of sparse representations, recent methods employ feature diffusion \cite{SAFDNet}, hierarchical aggregation \cite{HEDNet}, sparse transformers \cite{DSVT}, and linear-complexity architectures \cite{VoxelMamba,lion}.
Camera-based methods face different challenges at long range, as depth uncertainty increases and small-object observations degrade with distance. 
Far3D~\cite{Far3D} constructs long-range queries from 2D priors, and LR3D~\cite{LR3D} leverages 2D box supervision to compensate for the lack of 3D annotations at long distances.

Existing multimodal long-range fusion work has mainly focused on LiDAR-camera settings.
SparseFusion \cite{sparsefusion} aligns object-level features in 3D space with lightweight attention, and SparseLIF \cite{sparselif} constructs 3D queries from image-based geometric cues with uncertainty-aware fusion. 
Self-supervised approaches for learning sparse fusion representations have also been explored \cite{Self-Supervised_Sparse}. 
However, these methods rely on relatively accurate LiDAR geometry for proposal generation and alignment, making them difficult to extend to 4D radar-camera settings where observations are substantially sparser and noisier.

\begin{figure}[t]
	\centering
        \centerline{\includegraphics[width=0.99\textwidth]{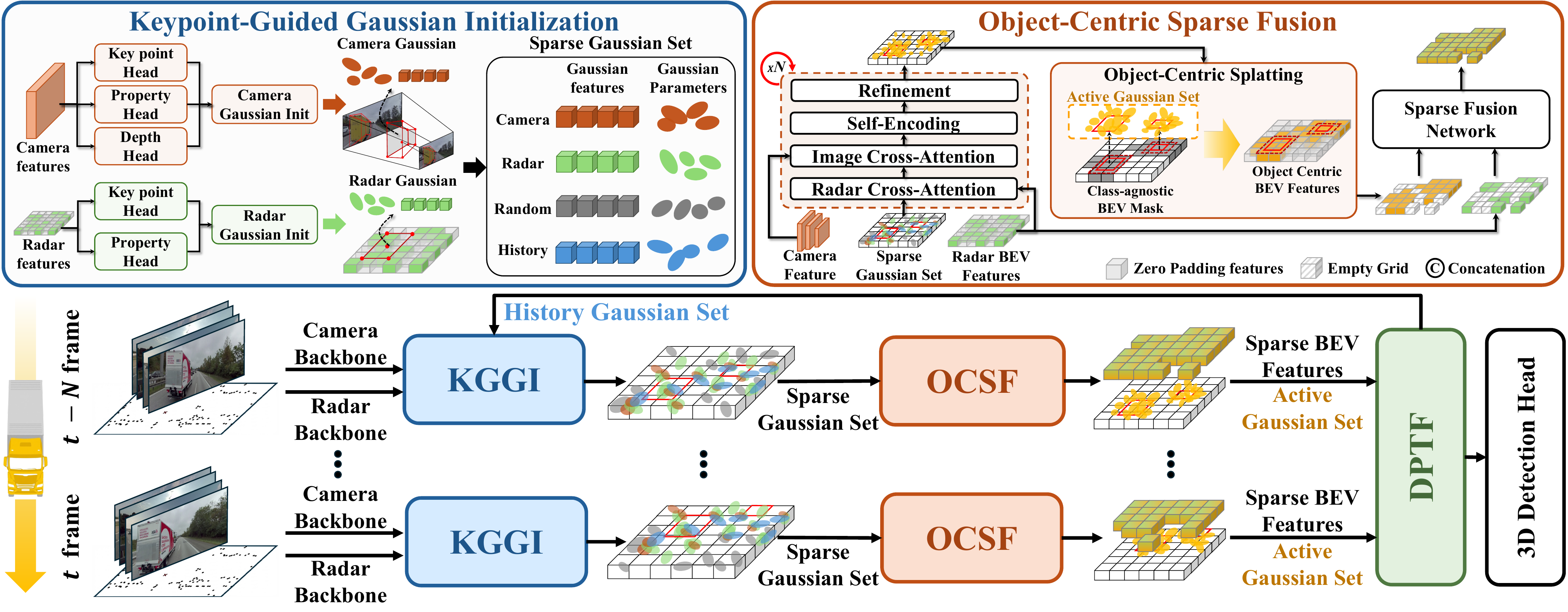}}
        \caption {\textbf{Overall architecture of Horizon3D.} Multi-view camera images and 4D radar points are encoded by  their respective backbones. The KGGI module estimates object keypoints from both modalities and initializes sparse Gaussian primitives at these locations. The OCSF module aggregates cross-modal features around the Gaussians and produces a sparse BEV representation that captures both object-level detail and scene-level context. The DPTF module then fuses temporal information through two complementary paths—Gaussian and BEV—to produce the fused feature map, which is passed to a 3D detection head.}
	\label{fig:overall}
\end{figure}

\section{Method}
\begin{figure}[t]
	\centering
        \centerline{\includegraphics[width=0.99\textwidth]{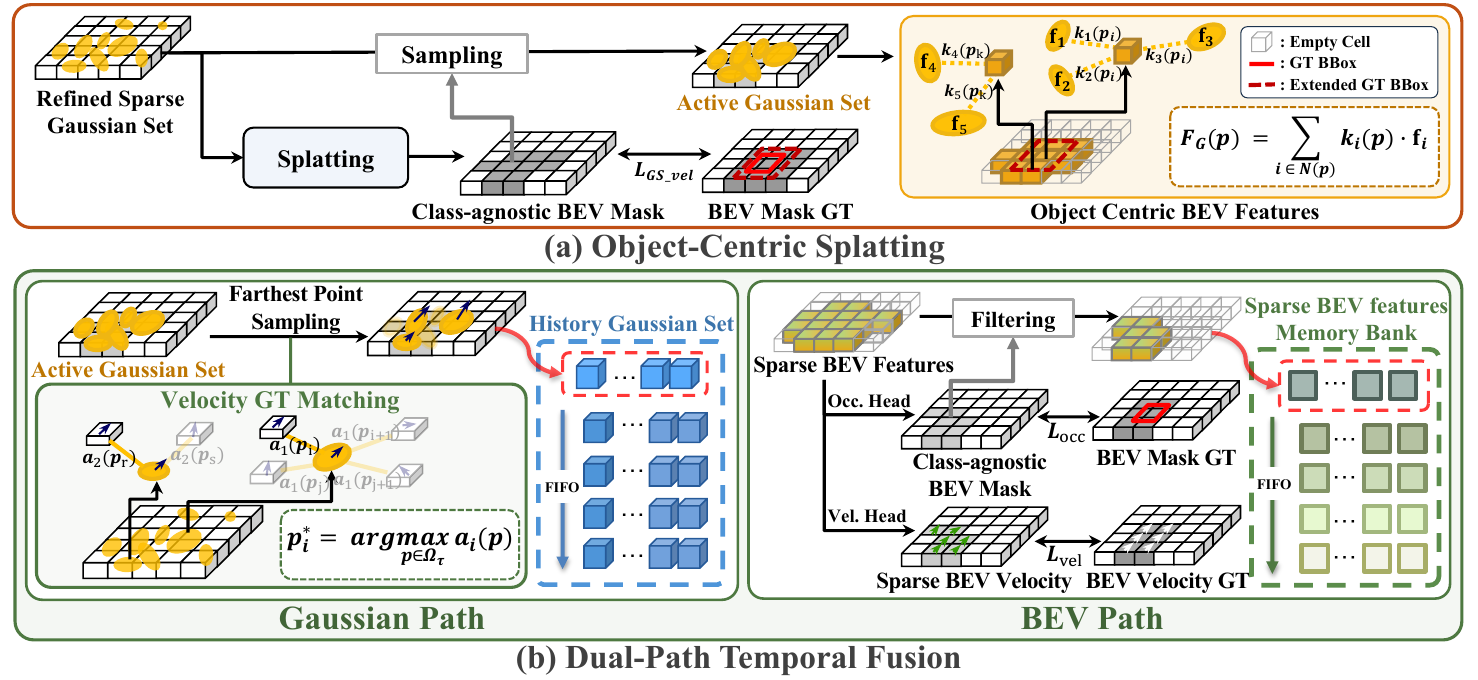}}
        \caption {\textbf{Details of Object-Centric Splatting and Velocity-Guided Temporal Alignment} 
        (a) Refined Gaussians are splatted onto BEV cells to produce a supervised occupancy mask. The Active Gaussian Set is pooled via Gaussian-weighted aggregation into object-centric BEV features.
        (b) For the Gaussian path (left), the Active Gaussian Set is subsampled via FPS to form the History Gaussian Set, with per-Gaussian velocity supervised by the most-contributed occupied cell. Stored Gaussians are warped via predicted velocities and ego-motion, then fed into the next frame's KGGI. For the BEV path (right), occupancy and a velocity head predict per-cell occupancy and velocity from the fused sparse BEV features, selecting foreground cells for memory storage and velocity-guided temporal alignment.}
	\label{fig:Fig3}
\end{figure}
\label{main_sec:method}
The overall architecture of Horizon3D is illustrated in Fig.~\ref{fig:overall}. Horizon3D consists of three key modules: \textbf{Keypoint-Guided Gaussian Initialization (KGGI)}, \textbf{Object-Centric Sparse Fusion (OCSF)}, and \textbf{Dual-Path Temporal Fusion (DPTF)}. 
Section~\ref{sec:KGGI} introduces KGGI, which initializes sparse Gaussian primitives at object keypoints predicted from radar and camera features.
Section~\ref{sec:OCSF} presents OCSF, which iteratively aggregates cross-modal features and refines the Gaussians to produce an object-centric BEV representation.
Section~\ref{sec:DPTF} describes DPTF, which jointly aggregates temporal cues through a BEV path that accumulates multi-frame features and a Gaussian path that propagates primitives across frames to encode per-object motion.

\subsection{Keypoint-Guided Gaussian Initialization}
\label{sec:KGGI}
Gaussian primitives have been explored as a flexible scene representation for 3D perception~\cite{GaussianFormer, GaussianFormerV2, GaussianFusion, RaGS}. 
These methods typically maintain a large set of Gaussians across the scene and iteratively refine their parameters to concentrate primitives in foreground regions. 
However, sufficient spatial coverage requires a substantial number of primitives, and the computational cost scales with the detection range.
To address this limitation, we propose Keypoint-Guided Gaussian Initialization, which estimates object keypoints from radar and camera features and anchors a compact set of Gaussians at these locations, enabling object-centric feature aggregation over a wide detection range without redundant spatial coverage.

We process 4D radar points, which provide spatial measurements, radar cross-section (RCS), and radial velocity, using VoxelNeXt~\cite{voxelnext}.
VoxelNeXt maintains a fully sparse representation while capturing broad scene-level context through its large receptive field, producing sparse radar BEV features $F_R \in \mathbb{R}^{C \times H_R \times W_R}$, where $C$, $H_R$, and $W_R$ denote the channel dimension, BEV height, and BEV width, respectively.
Given $N$ surround-view camera images, we extract image features $F_{I} \in \mathbb{R}^{N \times C_I \times H_I \times W_I}$ using a camera backbone.

For the radar branch, we apply a lightweight MLP to the per-cell features in $F_R$ to predict objectness scores and select the top-$K_r$ scoring cells as radar keypoints. These keypoints are used to initialize $K_r$ Gaussian primitives. 
For the camera branch, inspired by Far3D~\cite{Far3D}, we feed $F_I$ into a 2D keypoint head to produce a per-pixel objectness score map $S_n \in \mathbb{R}^{1 \times H_I \times W_I}$ for each view.
In parallel, a depth head predicts a depth distribution $D_n \in \mathbb{R}^{N_d \times H_I \times W_I}$, where $N_d$ is the number of depth bins.
At each pixel $(u, v)$ in view $n$, we obtain the estimated depth $d$ and its confidence from $D_n$, and compute a candidate score by combining the objectness score $S_n(u,v)$ with the depth confidence.
The top-$K_c$ scoring candidates are then lifted to 3D using the estimated depth and camera calibration and projected onto the BEV plane as camera keypoints. These keypoints are also used for $K_c$ Gaussian primitives.

Each Gaussian is parameterized as
\begin{equation}
\mathbf{g}_i = (\mathbf{m}_i \in \mathbb{R}^2,\ \mathbf{s}_i \in \mathbb{R}^2,\ \theta_i \in \mathbb{R},\ o_i \in \mathbb{R},\ \mathbf{v}_i \in \mathbb{R}^2),
\end{equation}
where $\mathbf{m}_i$, $\mathbf{s}_i$, $\theta_i$, $o_i$, and $\mathbf{v}_i$ denote the mean, scale, rotation, opacity, and velocity of the $i$-th Gaussian, respectively. 
Each Gaussian is also associated with a feature vector $\mathbf{f}_i \in \mathbb{R}^C$. The $K_r$ and $K_c$ Gaussians are initialized using the radar key points and camera key points, respectively.
Specifically, we initialize $\mathbf{m}_i$ with the corresponding keypoint location. The remaining attributes $(\mathbf{s}_i,\theta_i,o_i, \mathbf{v}_i)$ and the feature $\mathbf{f}_i$ are initialized as learnable parameters, which are refined by the OCSF module.
In addition to $K_r+K_c$ Gaussians, we add $K_{\text{rand}}$ randomly initialized Gaussians with learnable parameters to improve coverage in regions not covered by either branch.
We also include $K_{\text{hist}}$ history Gaussians propagated from previous frames to provide object-level temporal cues for the current frame (Sec.~\ref{sec:DPTF}). 
The resulting sparse Gaussian set is defined as $\mathcal{G}_0=\{\mathbf{g}_i\}_{i=1}^{N_g}$, where $N_g = K_r + K_c + K_{\text{rand}} + K_{\text{hist}}$.

\subsection{Object-Centric Sparse Fusion}
\label{sec:OCSF}
As discussed in Sec.~\ref{sec:intro}, dense BEV-based methods incur prohibitive computational cost at extended ranges, whereas query-based methods provide limited scene-level context. To address these limitations, OCSF takes the initialized Gaussian set $\mathcal{G}_0$ and progressively aggregates cross-modal features around the Gaussian primitives. This process iteratively refines the Gaussian parameters, allowing the primitives to better align with the underlying 3D object geometry. The refined Gaussians are then splatted onto the BEV plane to generate a sparse object-centric BEV representation, which is fused with sparse radar BEV features to jointly capture fine-grained object-level information and scene-level context.

\noindent\textbf{Multimodal Gaussian Encoder.} 
Motivated by recent progress in Gaussian-based representations \cite{GaussianFormer, GaussianFormerV2}, we propose a Multimodal Gaussian Encoder that refines the initialized primitives over $L$ layers, where each layer updates both Gaussian features and parameters.
At layer $l$, each Gaussian gathers modality-specific contexts through deformable cross-attention~\cite{DeformableDETR}. 
For each Gaussian, its mean $\mathbf{m}_i$ serves as the reference point.
Around this reference point, we generate $K$ sampling points $\mathcal{Q}_i = \{\mathbf{m}_i + \boldsymbol{\Delta}_k\}_{k=1}^{K}$.
The offsets $\boldsymbol{\Delta}_k$ are derived from the scale $\mathbf{s}_i$ and rotation $\theta_i$ to distribute the sampling points along the Gaussian's spatial extent.
Additional learnable offsets predicted from $\mathbf{f}_i$ further adapt the sampling locations to regions of interest.
The Gaussian feature is then updated as
\begin{equation}
\hat{\mathbf{f}}_i^{(l)} = \mathbf{f}_i^{(l)} + \sum_{M \in \{R, I\}} 
\mathrm{DCA}_M\!\left(\mathbf{f}_i^{(l)},\, F_M,\, \phi_M \right),
\label{eq:dac}
\end{equation}
where $\mathrm{DCA}_M$ is deformable cross-attention over modality $M$, and $F_R$ and $F_I$ are the radar BEV and multi-view image features, respectively.
The projection $\phi_M$ maps the sampling points $\mathcal{Q}_i$ to feature coordinates in modality $M$, where the projected locations serve as sampling locations for deformable cross-attention.
The projection $\phi_M$ maps the sampling points $\mathcal{Q}_i$ to modality-specific feature coordinates used by deformable cross-attention.
Specifically, $\phi_R$ maps the 2D sampling points to BEV cell coordinates for radar feature sampling.
For image feature sampling, each 2D sampling point in $\mathcal{Q}_i$ is lifted from the BEV plane to 3D by predicting an additional height coordinate from $\mathbf{f}_i$, and $\phi_I$ projects the resulting 3D points onto each camera view.

We quantize the 2D mean $\mathbf{m}_i$ of each Gaussian onto the BEV grid and apply 2D sparse convolution over the occupied BEV cells, enabling interactions among nearby Gaussians.
The mean is refined as
\begin{equation}
\mathbf{m}_i^{(l+1)} = \mathbf{m}_i^{(l)} + \mathbf{r}_i^{(l)}, \quad \mathbf{r}_i^{(l)} = \mathrm{MLP}_m\!\left(\hat{\mathbf{f}}_i^{(l)}\right),
\label{eq:refine}
\end{equation}
where $\mathbf{r}_i^{(l)}$ is the predicted residual. 
The remaining parameters $(\mathbf{s}_i,\,\theta_i,\,o_i)$ and per-Gaussian velocity $\mathbf{v}_i$ for temporal fusion (Sec.~\ref{sec:DPTF}) are also regressed from $\hat{\mathbf{f}}_i^{(l)}$ at each layer. 
The output feature after sparse convolution is used as $\mathbf{f}_i^{(l+1)}$ and passed to the next layer. After $L$ layers, we obtain the refined Gaussian set $\mathcal{G}_L$.

\noindent\textbf{Object-Centric Gaussian Splatting.}
The refined Gaussian set $\mathcal{G}_L$ is splatted onto BEV cells to produce an object-centric BEV representation (Fig.~\ref{fig:Fig3}(a)).
For each Gaussian $\mathbf{g}_i$, we construct its 2D covariance matrix as
\begin{equation}
\Sigma_i = R(\theta_i)\,\mathrm{diag}(s_{ix}^2,\,s_{iy}^2)\,R(\theta_i)^{\top},
\end{equation}
where $R(\theta_i)$ is the 2D rotation matrix.
At a BEV cell center $\mathbf{p}$, the Gaussian kernel weight is given by $k_i(\mathbf{p})=\exp\!\left(-\tfrac{1}{2}(\mathbf{p}-\mathbf{m}_i)^{\top}\Sigma_i^{-1}(\mathbf{p}-\mathbf{m}_i)\right)$, and the occupancy contribution is given by $a_i(\mathbf{p})=\sigma(o_i)\,k_i(\mathbf{p})$.
The class-agnostic occupancy probability at $\mathbf{p}$ is then
\begin{equation}
\alpha(\mathbf{p})=1-\prod_{i=1}^{N_g}\bigl(1-a_i(\mathbf{p})\bigr).
\end{equation}
The resulting occupancy map $\alpha \in \mathbb{R}^{H_R \times W_R}$ is supervised using a ground-truth BEV mask derived from rasterized 3D bounding boxes, with each box enlarged by a margin $\epsilon$ to encourage encoding of surrounding spatial context.

BEV cells exceeding an occupancy threshold, $\Omega_\tau=\{\mathbf{p}\mid \alpha(\mathbf{p})>\tau\}$, form a sparse object-centric BEV representation. 
For each selected cell, we define the Active Gaussian Set $\mathcal{N}(\mathbf{p})$ as the set of Gaussians contributing to that cell, and compute its BEV feature by Gaussian-weighted pooling as
\begin{equation}
F_G(\mathbf{p})=\sum_{i\in\mathcal{N}(\mathbf{p})} k_i(\mathbf{p})\,\mathbf{f}_i,\qquad \mathbf{p}\in\Omega_\tau.
\end{equation}
The object-centric BEV features $F_G$ are then fused with the sparse radar BEV features $F_R$ to produce the final sparse BEV representation. 
We take the union of non-empty BEV cells in $F_G$ and $F_R$ and apply a lightweight gating network to adaptively weight both features. 
The gated features are processed by a sparse encoder with submanifold sparse convolutions, yielding the fused sparse BEV feature map $F_{\mathrm{fuse}}$. 
We provide the full formulation in the supplementary material.

\subsection{Dual-Path Temporal Fusion}
\label{sec:DPTF}
As discussed in Sec.~\ref{sec:intro}, long-range objects produce sparse per-frame observations that demand multi-frame accumulation, while fast-moving objects undergo large inter-frame displacements that require object-level motion modeling.
To address both challenges, we propose DPTF, which integrates temporal cues through two complementary paths. The BEV path compensates for sparsity by accumulating BEV features over multiple frames, while operating only on object-occupied cells rather than on the entire dense grid. 
The Gaussian path warps past Gaussians to the current frame and merges them with current-frame primitives to capture object-level temporal dynamics. 
In both paths, predicted velocities are used alongside ego-motion to handle large displacements of fast-moving objects.

\noindent\textbf{BEV Path.}
The BEV path accumulates foreground BEV features over multiple frames to compensate for the sparse observations of distant objects (Fig.~\ref{fig:Fig3}(b), right). 
Given the fused sparse BEV feature map $F_{\mathrm{fuse}}^{t}$, we predict an occupancy score $\hat{o}^{t}(\mathbf{p})$ and a planar velocity vector $\hat{\mathbf{u}}^{t}(\mathbf{p})$ at each non-empty BEV cell $\mathbf{p}$ as
\begin{equation}
\hat{o}^{t}(\mathbf{p})=\sigma\!\left(h_{\mathrm{occ}}\!\left(F_{\mathrm{fuse}}^{t}(\mathbf{p})\right)\right),\qquad
\hat{\mathbf{u}}^{t}(\mathbf{p})=h_{\mathrm{vel}}\!\left(F_{\mathrm{fuse}}^{t}(\mathbf{p})\right),
\end{equation}
where $h_{\mathrm{occ}}$ and $h_{\mathrm{vel}}$ are small stacks of submanifold sparse convolutions.
While the occupancy mask in Sec.~\ref{sec:OCSF} uses enlarged bounding boxes to encourage contextual encoding, $\hat{o}^{t}(\mathbf{p})$ is supervised with a tight mask from the original boxes, yielding a compact set of object-occupied cells for efficient memory storage and temporal aggregation.
The predicted velocity $\hat{\mathbf{u}}^{t}(\mathbf{p})$ is supervised by per-cell ground-truth velocities, where each cell within a ground-truth bounding box is assigned the velocity of that box.
We select non-empty cells whose occupancy scores exceed a threshold $\tau_{\mathrm{mem}}$ as foreground cells and store only their features $F_{\mathrm{fuse}}^{t}(\mathbf{p})$ and predicted velocities $\hat{\mathbf{u}}^{t}(\mathbf{p})$ in the BEV memory bank.

For temporal fusion, we align each historical frame $(t{-}k)$ to the current frame by combining ego-motion with velocity-guided warping.
The warped BEV cell index is computed as 
\begin{equation}
\mathbf{p}'=\pi\!\left(T_{t-k\rightarrow t}\bigl(\mathbf{x}(\mathbf{p})+\hat{\mathbf{u}}^{t-k}(\mathbf{p})\,\Delta t_k\bigr)\right),
\end{equation}
where $\mathbf{p}'$ is the warped BEV cell index in the current frame, $T_{t-k\rightarrow t}$ is the ego-motion transformation from frame $t{-}k$ to $t$, $\Delta t_k$ is the elapsed time between the two frames, $\mathbf{x}(\cdot)$ maps a BEV cell center to world coordinates, and $\pi(\cdot)$ quantizes the warped coordinates back to BEV indices.
The aligned historical BEV features are stacked with the current fused BEV feature map $F_{\mathrm{fuse}}^{t}$ along the temporal axis.
Each temporal feature map is augmented with a time embedding derived from $\Delta t_k$, and sparse convolutions fuse the stacked features into a temporally fused sparse BEV feature map for the detection head.
We provide the detailed architecture of this aggregation module in the supplementary material.

\noindent\textbf{Gaussian Path.}
The Gaussian path captures object-level temporal dynamics by selecting and propagating past Gaussians to the current frame (Fig.~\ref{fig:Fig3}(b), left).
After splatting, we gather Gaussians that contribute to occupied cells via the Active Gaussian Set $\mathcal{N}(\mathbf{p})$ and apply farthest point sampling (FPS) to form a spatially diverse History Gaussian Set for memory storage.
Compared to top-$K$ selection by opacity, which tends to concentrate on a few high-response regions, FPS samples Gaussians more evenly across objects, improving object-level temporal modeling after merging with current-frame primitives.

At highway speeds, dynamic objects can move substantially between consecutive frames, and ego-motion compensation alone leaves significant spatial misalignment. 
The velocity $\mathbf{v}_i$ predicted by the OCSF refinement module (Sec.~\ref{sec:OCSF}) is supervised using the ground-truth velocity of the occupied cell to which each Gaussian contributes most.
The target cell and velocity are defined as
\begin{equation}
\mathbf{p}_i^{*} = \arg\max_{\mathbf{p}\in\Omega_\tau}\, a_i(\mathbf{p}), \qquad
\mathbf{v}_i^{\mathrm{gt}} = \mathbf{v}^{\mathrm{gt}}(\mathbf{p}_i^{*}).
\end{equation}
Each Gaussian's predicted velocity is then employed alongside ego-motion to warp past Gaussians to the current frame. 
The warped History Gaussian Set is incorporated into the sparse Gaussian set of KGGI (Sec.~\ref{sec:KGGI}) and jointly processed with current-frame Gaussians through the OCSF pipeline (Sec.~\ref{sec:OCSF}). 
Since residual misalignment can remain after velocity compensation, we add temporal self-attention within the Gaussian encoder over the merged set, enabling cross-frame interaction beyond the local receptive field of sparse convolution.

\section{Experiments}
\label{main_sec:exp}

\subsection{Experimental Setup}
\noindent{\bf Datasets and Metrics.}
We evaluate our proposed method on the TruckScenes~\cite{truckscenes} dataset. The TruckScenes dataset consists of 747 driving scenes, each approximately 20 seconds long, captured by four cameras, six LiDARs, and six 4D radars mounted on a heavy-duty commercial truck. Unlike existing urban-driving benchmarks, TruckScenes features 
high-speed driving scenarios with an evaluation range of up to 150m. Our evaluation follows the official benchmark metrics, including mean Average Precision (mAP) and nuScenes Detection Score (NDS).

\noindent{\bf Implementation Details.}
We employ VoVNet-99~\cite{vovnet} initialized from a FCOS3D ~\cite{fcos3D} checkpoint following SpaRC~\cite{SpaRC} and ResNet-50~\cite{resnet} pretrained on ImageNet as camera backbones. In the radar branch, we accumulate the past $10$ sweeps from six radar sensors as input point clouds and use VoxelNeXt~\cite{voxelnext} as the backbone network. Training proceeds in two phases: we first train the single-frame model (KGGI and OCSF) for $24$ epochs, then train the full model including DPTF for an additional $24$ epochs. For temporal fusion, the BEV path aggregates features from the past $8$ frames, while the Gaussian path maintains history Gaussians from the past $4$ frames. We use AdamW with an initial learning rate of $1{\times}10^{-4}$ and a weight decay of $1{\times}10^{-2}$. All experiments are conducted on NVIDIA RTX PRO 5000 Blackwell and RTX 3090 GPUs, and inference speed is measured on a single RTX 3090. Detailed training configurations and hyperparameter settings are provided in the supplementary material.
\begin{table}[t!]
\centering
\setlength{\tabcolsep}{2pt}
\fontsize{9pt}{10pt}\selectfont

\caption{Performance comparison on the TruckScenes dataset. ‘L’, ‘C’, and ‘R’ denote LiDAR, Camera, and Radar, respectively. * indicates results from the official TruckScenes baseline. \dag~indicates methods reproduced using officially released code 
following their original training configurations.}
\label{tab:main_comparison}
\vspace{-10pt}

\resizebox{\textwidth}{!}{ 
    \begin{tabular}{l|c|c|c|c|>{\hspace{0.05cm}}ccccccc}
    \toprule[1.2pt]
    \textbf{Methods} & \textbf{Input} & \textbf{Backbone} & \textbf{Img Size} & \textbf{Split} & \textbf{NDS$\uparrow$} & \textbf{mAP$\uparrow$} & \textbf{mATE$\downarrow$} & \textbf{mASE$\downarrow$} & \textbf{mAOE$\downarrow$} & \textbf{mAVE$\downarrow$} & \textbf{mAAE$\downarrow$} \\
    \midrule[0.4pt]

    CenterPoint-V* \cite{Centerpoint} & L & Voxel & -- & val & 35.3 & 22.6 & 0.461 & 0.405 & 0.468 & 3.028 & 0.261 \\
    RCTrans\dag  \cite{RCTrans}& C+R & R50 & 256$\times$864 & val & 22.9 & 12.6 & 1.025 & 0.507 & 0.608 & 0.894 & 0.331 \\
    CRT-Fusion\dag \cite{CRT-Fusion} & C+R & R50 & 256$\times$864 & val & 28.8 & 16.9 & 0.833 & 0.512 & 0.444 & 0.852 & 0.322 \\
    BEVFusion\dag \cite{BEVFusion}& C+R & R50 & 256$\times$864 & val & 30.4 & 18.2 & 0.941 & 0.442 & 0.389 & 0.892 & 0.208 \\
    \rowcolor[gray]{0.90}
    \textbf{Horizon3D (Ours)} & C+R & R50 & 256$\times$864 & val & \textbf{37.4} & \textbf{23.6} & 0.876 & 0.410 & 0.331 & 0.625 & 0.198 \\
    
    \midrule[0.4pt]
    Far3D \cite{Far3D} & C & V2-99 & 640$\times$960 & val & 21.4 & 10.7 & 0.883 & 0.507 & 0.671 & 1.352 & 0.338 \\
    SpaRC \cite{SpaRC}& C+R & V2-99 & 640$\times$960 & val & 35.4 & 22.5 & 0.798 & 0.449 & 0.476 & 0.613 & 0.248 \\
    \rowcolor[gray]{0.90}
    \textbf{Horizon3D (Ours)} & C+R & V2-99 & 640$\times$960 & val & \textbf{38.4} & \textbf{24.1} & 0.833 & 0.404 & 0.355 & 0.561 & 0.208 \\
    
    \midrule[0.4pt]

    CenterPoint-V* \cite{Centerpoint}& L & Voxel & -- & test & 41.0 & 26.7 & 0.409 & 0.352 & 0.277 & 2.730 & 0.201 \\
    RadarGNN* \cite{RadarGNN} & R & -- & -- & test & 10.7 & 7.0 & 0.892 & 0.809 & 1.132 & 8.003 & 0.571 \\
    PETR* \cite{PETR} & C & V2-99 & 300$\times$800 & test & 12.1 & 2.2 & 1.125 & 0.686 & 0.647 & 1.499 & 0.564 \\
    HyDRa \cite{HyDRa} & C+R & V2-99 & 928$\times$1952 & test & 22.4 & 12.8 & 0.725 & 0.544 & 0.744 & 1.180 & 0.388 \\
    SpaRC \cite{SpaRC}& C+R & V2-99 & 928$\times$1952 & test & 37.4 & 27.2 & 0.759 & 0.413 & 0.411 & 0.814 & 0.227 \\
    \rowcolor[gray]{0.90}
    \textbf{Horizon3D (Ours)}& C+R & V2-99 & 640$\times$960 & test & \textbf{41.8} & \textbf{27.7} & 0.779 & 0.354 & 0.269 & 0.637 & 0.167 \\

    \bottomrule[1.2pt] 
    \end{tabular}
}
\end{table}

\begin{table}[t!]
    \centering
    \begin{minipage}[t]{0.48\textwidth} 
\centering
\setlength{\tabcolsep}{2pt}
\scriptsize

\caption{Ablation study of main components of Horizon3D.}
\label{tab:ablation_main}
\vspace{-10pt}
\begin{tabular}{c|ccc|cc}
\toprule[1.2pt]
\textbf{Input} & \textbf{KGGI} & \textbf{OCSF} & \textbf{DPTF} & \textbf{mAP$\uparrow$} & \textbf{NDS$\uparrow$} \\
\midrule[0.4pt]
R & & & & 12.8 & 27.1 \\
\midrule[0.4pt]
\multirow{4}{*}{C+R}& & \checkmark & & 17.7 & 32.5 \\
& \checkmark & \checkmark & & 20.3 & 35.4 \\
& & \checkmark & \checkmark & 23.0 & 36.4 \\
& \checkmark & \checkmark & \checkmark & \textbf{23.6} & \textbf{37.4} \\ 
\bottomrule[1.2pt] 
\end{tabular}
\end{minipage}
    \hfill
    \begin{minipage}[t]{0.48\textwidth} 
\centering
\setlength{\tabcolsep}{5pt}
\scriptsize

\caption{Ablation study of Gaussian initialization strategies in KGGI.}
\label{tab:ablation_kggi_2}
\vspace{-10pt}
\begin{tabular}{c|>{\hspace{0.05cm}}cc}
\toprule[1.2pt]
 & \textbf{mAP$\uparrow$} & \textbf{NDS$\uparrow$} \\
\midrule[0.4pt]
Random & 17.7 & 32.5 \\
Keypoint only & 19.2 & 34.3 \\
\rowcolor[gray]{0.90}
\midrule[0.4pt]
\textbf{Random + Keypoint} & \textbf{20.3} & \textbf{35.4} \\ 
\bottomrule[1.2pt] 
\end{tabular}
\end{minipage}
\end{table}

\subsection{Comparison to the state of the art}
Table~\ref{tab:main_comparison} compares Horizon3D with existing methods on the TruckScenes dataset. 
On the validation split with a ResNet-50 backbone, Horizon3D achieves the highest performance among radar-camera methods, outperforming both BEV-based methods~\cite{BEVFusion,CRT-Fusion} and the query-based method~\cite{RCTrans} by a large margin.
Notably, Horizon3D also surpasses CenterPoint-V~\cite{Centerpoint}, a LiDAR-based method, by $\mathbf{+2.1}$ NDS and $\mathbf{+1.0}$ mAP.
With the V2-99 backbone, Horizon3D exceeds SpaRC~\cite{SpaRC}, the previous state-of-the-art radar-camera method, by $\mathbf{+3.0}$ NDS and $\mathbf{+1.6}$ mAP.
On the test split, Horizon3D further outperforms SpaRC by $\mathbf{+4.4}$ NDS and $\mathbf{+0.5}$ mAP while using lower-resolution image inputs, demonstrating the effectiveness of our hybrid representation.
Horizon3D also surpasses CenterPoint-V~\cite{Centerpoint} by $\mathbf{+0.8}$ NDS and $\mathbf{+1.0}$ mAP, achieving competitive performance against LiDAR-based methods.

\begin{table}[t!]
    \centering
    \begin{minipage}[t]{0.48\textwidth} 
\centering
\setlength{\tabcolsep}{8pt}
\scriptsize

\caption{Ablation study of bounding box enlargement for occupancy supervision.}
\label{tab:OCSF_ablation_1}
\vspace{-10pt}
\begin{tabular}{c|>{\hspace{0.05cm}}cc}
\toprule[1.2pt]
\textbf{Enlarged GT} & \textbf{mAP$\uparrow$} & \textbf{NDS$\uparrow$} \\
\midrule[0.4pt]
 & 17.2 & 32.9 \\
\rowcolor[gray]{0.90}
\midrule[0.4pt]
\checkmark & \textbf{20.3} & \textbf{35.4} \\ 
\bottomrule[1.2pt] 
\end{tabular}
\end{minipage}

    \hfill
    \begin{minipage}[t]{0.48\textwidth} 
\centering
\setlength{\tabcolsep}{4pt}
\scriptsize

\caption{Ablation study for the BEV and Gaussian paths in DPTF.}
\label{tab:DPTF_ablation_1}
\vspace{-10pt}
\begin{tabular}{cc|>{\hspace{0.05cm}}cc}
\toprule[1.2pt]
\hspace{5pt}\textbf{BEV}\hspace{5pt} & \hspace{5pt}\textbf{Gaussian}\hspace{5pt} & \textbf{mAP$\uparrow$} & \textbf{NDS$\uparrow$} \\
\midrule[0.4pt]
\checkmark & & 22.5 & 36.3 \\
& \checkmark & 21.3 & 36.0 \\
\rowcolor[gray]{0.90}
\midrule[0.4pt]
\checkmark & \checkmark & \textbf{23.6} & \textbf{37.4} \\ 
\bottomrule[1.2pt] 
\end{tabular}
\end{minipage}

\end{table}

\subsection{Ablation Studies}
We conduct ablation studies on the TruckScenes validation set using the ResNet-50 backbone. 
For variants without DPTF, we report results from the first training phase (single-frame model) for efficiency.

\noindent\textbf{Component analysis.}
To assess the contribution of each module, we incrementally add components and report the results in Table~\ref{tab:ablation_main}. 
The radar-only baseline achieves 12.8 mAP and 27.1 NDS.
Adding OCSF with randomly initialized Gaussians improves mAP by $+4.9$ and NDS by $+5.4$, showing that Gaussian-based cross-modal fusion effectively injects object-level detail into the BEV space.
Replacing random initialization with KGGI further improves mAP by $+2.6$ and NDS by $+2.9$, demonstrating that placing Gaussians at estimated object keypoints leads to more effective object-centric encoding. 
Adding DPTF without KGGI improves over the OCSF-only variant by $+5.3$ mAP and $+3.9$ NDS, showing the effectiveness of dual-path temporal fusion.
Combining all three modules achieves the best performance, with an mAP of 23.6 and an NDS of 37.4, confirming that the modules provide complementary benefits.

\noindent\textbf{Initialization strategies for KGGI.}
Table~\ref{tab:ablation_kggi_2} compares Gaussian initialization strategies. 
Keypoint-only initialization outperforms random initialization by $+1.5$ mAP and $+1.8$ NDS, confirming the importance of sensor-guided placement. 
Combining keypoint and random Gaussians further improves both metrics by $+1.1$, as random Gaussians help cover regions missed by sensor-guided keypoints.

\noindent\textbf{Effect of Bounding Box Enlargement.}
Table~\ref{tab:OCSF_ablation_1} examines the effect of bounding box enlargement for occupancy supervision in OCSF (Sec.~\ref{sec:OCSF}). 
Without enlargement, the model achieves 17.2 mAP and 32.9 NDS.
Enlarging the bounding boxes by a margin $\epsilon$ before rasterization improves mAP by $+3.1$ and NDS by $+2.5$, as the expanded supervision encourages Gaussians to encode not only object interiors but also surrounding spatial context.

\begin{table}[t!]
    \centering
    \begin{minipage}[t]{0.48\textwidth} 
\scriptsize
\centering
\setlength{\tabcolsep}{2pt}
\renewcommand{\arraystretch}{0.7}
\caption{Ablation study for evaluating the effect of object velocity compensation.}
\label{tab:DPTF_ablation_2}
\vspace{-10pt}
\resizebox{\textwidth}{!}{
\begin{tabular}{c|c|cc}
\toprule[1.2pt]
\textbf{Path} & \makecell{\textbf{Object Vel.} \\ \textbf{Compensation}} & \textbf{mAP$\uparrow$} & \textbf{NDS$\uparrow$} \\
\midrule[0.4pt]
\multirow{2}{*}{\textbf{BEV}} & & 21.6 & 36.1 \\
& \checkmark & \textbf{23.0} & \textbf{36.7} \\
\midrule[0.4pt]
\multirow{2}{*}{\textbf{Gaussian}} & & 20.6 & 34.8 \\
& \checkmark & \textbf{21.3} & \textbf{36.0} \\
\bottomrule[1.2pt]
\end{tabular}%
}
\end{minipage}
    \hfill
    \begin{minipage}[t]{0.48\textwidth} 
\centering
\setlength{\tabcolsep}{7pt}
\scriptsize

\caption{Ablation study on sampling strategies for Gaussian path.}
\label{tab:DPTF_ablation_3}
\vspace{-10pt}
\begin{tabular}{c|cc}
\toprule[1.2pt]
& \hspace{5pt}\textbf{mAP$\uparrow$}\hspace{5pt} & \hspace{5pt}\textbf{NDS$\uparrow$}\hspace{5pt} \\
\midrule[0.4pt]
Random & 20.7 & 35.8 \\
Top-K & 21.1 & 35.9 \\
\rowcolor[gray]{0.90}
\midrule[0.4pt]
\textbf{FPS} & \textbf{21.3} & \textbf{36.0}  \\
\bottomrule[1.2pt] 
\end{tabular}
\end{minipage}

\end{table}

\begin{table}[t!]
    \centering
    \begin{minipage}[t]{0.48\textwidth} 
\centering
\scriptsize
\caption{Comparison of accuracy and efficiency with existing radar–camera fusion methods.}
\label{tab:memory_fps}
\vspace{-10pt}
\resizebox{\textwidth}{!}{
\begin{tabular}{c|cc|cc}
\toprule[1.2pt]
\textbf{Method} & \textbf{mAP$\uparrow$} & \textbf{NDS$\uparrow$} & \textbf{FPS$\uparrow$} & \textbf{Mem(GB)$\downarrow$} \\
\midrule[0.4pt]
 BEVFusion \cite{BEVFusion} & 18.2 & 30.4 & 4.2 & 6.520\\
 CRTFusion \cite{CRT-Fusion} & 16.9 & 28.8 & 2.9 & 7.580 \\
 RCTrans \cite{RCTrans} & 12.6 & 22.9 & \textbf{9.5} & \textbf{2.620}\\
\rowcolor[gray]{0.90}
\midrule[0.4pt]
\textbf{Horizon3D} & \textbf{23.6} & \textbf{37.4} & 8.5 & 3.236 \\ 
\bottomrule[1.2pt] 
\end{tabular}
}
\end{minipage}
    \hfill
    \begin{minipage}[t]{0.48\textwidth} 
\centering
\scriptsize
\caption{Ablation study on distance-wise performance.}
\label{tab:distance}
\vspace{-5pt}

\resizebox{\columnwidth}{!}{
    \setlength{\tabcolsep}{2pt} 
    \begin{tabular}{l|cc|cc|cc|cc} 
    \toprule[1.2pt]
    & \multicolumn{2}{c|}{\textbf{0-25m}} & \multicolumn{2}{c|}{\textbf{25-50m}} & \multicolumn{2}{c|}{\textbf{50-100m}} & \multicolumn{2}{c}{\textbf{100-150m}} \\
    \cmidrule(lr){2-3} \cmidrule(lr){4-5} \cmidrule(lr){6-7} \cmidrule(lr){8-9}
    & mAP & NDS & mAP & NDS & mAP & NDS & mAP & NDS \\ 
    \midrule[0.4pt]
    RCTrans \cite{RCTrans} & 19.9 & 28.4 & 12.5 & 24.1 & 7.9 & 19.9 & 6.2 & 15.6 \\
    CRTFusion \cite{CRT-Fusion} & 26.2 & 34.4 & 16.5 & 27.3 & 11.0 & 26.3 & 7.0 & 11.4 \\
    BEVFusion  \cite{BEVFusion} & 28.8 & 35.7 & 16.8 & 25.2 & 11.7 & 20.1 & 7.4 & 13.9 \\
    \rowcolor[gray]{0.90}
    \midrule[0.4pt]
    \textbf{Horizon3D}  & \textbf{36.5} & \textbf{43.0} & \textbf{23.3} & \textbf{37.8} & \textbf{17.0} & \textbf{32.7} & \textbf{8.9} & \textbf{18.4} \\
    \bottomrule[1.2pt] 
    \end{tabular}
}
\end{minipage}
\end{table}

\noindent\textbf{Effect of Dual-Path Temporal Fusion.}
Tables~\ref{tab:DPTF_ablation_1} and \ref{tab:DPTF_ablation_2} analyze the design choices in DPTF (Sec.~\ref{sec:DPTF}).
In Table~\ref{tab:DPTF_ablation_1}, the BEV path alone achieves 22.5 mAP and 36.3 NDS, while the Gaussian path alone yields 21.3 mAP and 36.0 NDS.
Combining both paths improves over the BEV-only variant by $+1.1$ mAP and $+1.1$ NDS, demonstrating that object-level and scene-level temporal fusion provide complementary benefits. 
Table~\ref{tab:DPTF_ablation_2} evaluates the effect of object velocity compensation in each path.
Applying velocity compensation improves the BEV path by $+1.4$ mAP and $+0.6$ NDS, and the Gaussian path  by $+0.7$ mAP and $+1.2$ NDS, confirming that explicit motion compensation is essential for both paths in high-speed scenarios.

\begin{figure}[t]
	\centering
        \centerline{\includegraphics[width=0.95\textwidth]{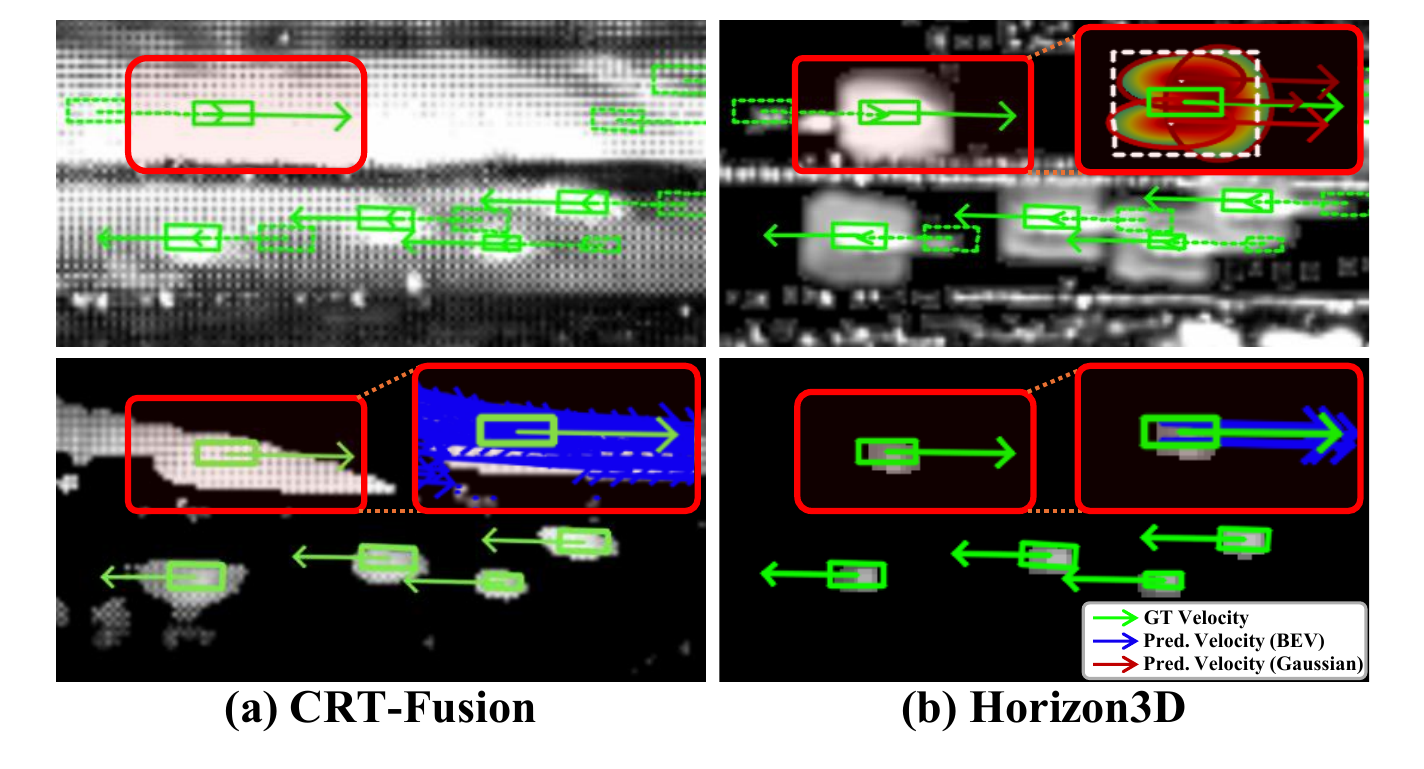}}
        \caption{\textbf{Qualitative comparison of temporally aggregated BEV feature maps (top) and velocity predictions with foreground regions (bottom).}
        Green and blue arrows denote ground-truth and predicted per-cell velocities, respectively. Red arrows indicate per-Gaussian velocities predicted by the Gaussian path. Dashed green boxes represent object positions from previous frames.}
	\label{fig:qual}
\end{figure}

\noindent\textbf{Effect of Gaussian Sampling Strategy.}
Table~\ref{tab:DPTF_ablation_3} compares strategies for selecting Gaussians to store in the temporal memory bank. 
Random selection achieves 20.7 mAP and 35.8 NDS, top-$K$ selection by opacity reaches 21.1 mAP and 35.9 NDS, and farthest point sampling (FPS) performs best with 21.3 mAP and 36.0 NDS. 
Top-$K$ selection tends to store Gaussians from a few high-response regions, whereas FPS selects a spatially diverse subset that better covers multiple objects, leading to more consistent temporal fusion after merging.

\noindent\textbf{Efficiency comparison.}
Table~\ref{tab:memory_fps} compares accuracy and efficiency with existing radar-camera fusion methods. 
Dense BEV-based methods \cite{BEVFusion, CRT-Fusion} suffer from high memory consumption and low inference speed due to the cost of dense BEV construction at extended ranges. 
RCTrans \cite{RCTrans} runs efficiently through query-based encoding but yields substantially lower detection accuracy. 
Horizon3D achieves the highest accuracy while maintaining competitive speed and moderate memory usage, demonstrating that the hybrid Gaussian-BEV representation effectively balances accuracy and efficiency for long-range perception.

\noindent\textbf{Distance-wise performance.} 
Table~\ref{tab:distance} reports performance across different distance ranges. 
Horizon3D achieves the highest mAP at every range, with particularly notable gains at $25$-$50$\,m ($+6.5$ mAP over BEVFusion) and $100$-$150$\,m ($+1.5$ mAP and $+4.5$ NDS over BEVfusion), confirming the effectiveness of our approach for long-range detection.

\noindent\textbf{Qualitative results.}
Fig.~\ref{fig:qual} compares CRT-Fusion and Horizon3D in terms of temporally aggregated BEV feature maps (top) and velocity predictions with foreground regions (bottom). 
CRT-Fusion shows widespread activations across the BEV grid, as temporal misalignment is further diffused by dense convolution. 
As shown in the bottom-left, foreground regions are inaccurately predicted and velocities are estimated over broad areas including background cells, introducing erroneous motion compensation. 
In contrast, Horizon3D preserves activations only around object regions (top-right). 
The object-centric Gaussian representation enables accurate foreground and velocity prediction in the fused BEV features (bottom-right), resulting in precise temporal alignment. 
The top-right zoom-in shows that Gaussians are distributed along the enlarged bounding box regions, extending activations slightly beyond object boundaries to capture surrounding context.

\section{Conclusion}
We presented Horizon3D, a sparse radar-camera fusion framework for long-range 3D object detection that captures both fine-grained object detail and scene-level context through a hybrid representation combining Gaussian primitives and sparse BEV features.
KGGI initializes a compact set of Gaussians at sensor-estimated object keypoints, and OCSF iteratively refines them to produce a sparse, object-centric BEV representation that captures both fine-grained object detail and scene-level context without dense BEV construction.
DPTF integrates temporal information through dual paths with explicit velocity compensation, maintaining both object-level dynamics and scene-level temporal context in high-speed scenarios.
Experiments on TruckScenes demonstrate that Horizon3D outperforms prior radar-camera fusion methods by $\mathbf{+3.0}$ NDS and $\mathbf{+1.6}$ mAP while maintaining competitive inference speed.




%
%
\bibliographystyle{splncs04}
\bibliography{main}
\clearpage
\begingroup

\setcounter{section}{0}
\setcounter{subsection}{0}

\renewcommand{\thesection}{\Alph{section}}
\renewcommand{\thesubsection}{\thesection.\arabic{subsection}}

\renewcommand{\theHsection}{supp.\Alph{section}}
\renewcommand{\theHsubsection}{supp.\Alph{section}.\arabic{subsection}}

\begin{center}
{\Large\bfseries Supplementary Material for\\[0.3em]
Horizon3D: Sparse Radar-Camera Fusion for Long-Range 3D Perception in Autonomous Driving}\\[0.8em]
\end{center}
\vspace{1em}

This supplementary material provides additional details that complement the main paper. 
We first describe the loss functions and network architecture details (Sections~\ref{sec:supp_loss}--\ref{sec:supp_arch}), followed by implementation details, including hyperparameters, data augmentation, and training configuration (Section~\ref{sec:supp_impl}). 
We then present additional experimental results with per-class analysis and ablation studies (Section~\ref{sec:supp_exp}).
Finally, we provide qualitative results and failure cases in Section~\ref{sec:supp_qual}.

\section{Loss Functions}
\label{sec:supp_loss}
The total training objective is defined as
\begin{equation}
    \mathcal{L}_{\text{total}} = \mathcal{L}_{\text{det}} + \lambda_{\text{kggi}}\mathcal{L}_{\text{kggi}} + \lambda_{\text{ocsf}}\mathcal{L}_{\text{ocsf}} + \lambda_{\text{dptf}}\mathcal{L}_{\text{dptf}},
\end{equation}
where $\mathcal{L}_{\text{det}}$ denotes the 3D detection loss, which combines a focal loss for classification and an L1 loss for bounding box regression following CenterPoint~\cite{Centerpoint}.
The KGGI loss $\mathcal{L}_{\text{kggi}}$ consists of a focal loss for 2D keypoint objectness prediction and a binary cross-entropy loss for depth distribution estimation following Far3D~\cite{Far3D}.
The OCSF loss $\mathcal{L}_{\text{ocsf}}$ applies focal and Dice losses to the class-agnostic occupancy map $\alpha(\mathbf{p})$ produced by Gaussian splatting (Eq.~5 of the main paper).
The ground-truth BEV mask is derived from enlarged bounding boxes, encouraging Gaussians to encode surrounding spatial context beyond object interiors.

The DPTF loss $\mathcal{L}_{\text{dptf}}$ combines losses from both temporal paths.
For the BEV path, per-cell occupancy is supervised with a tight BEV mask using focal, Dice, and Lov\'{a}sz-Softmax losses, while per-cell velocity is supervised with an L1 loss.
For the Gaussian path, we apply a Smooth L1 loss to per-Gaussian velocity, where each Gaussian is matched to the ground-truth velocity of the occupied cell receiving its largest contribution $\mathbf{p}_i^*$ (Eq.~9 of the main paper).
We set $\lambda_{\text{kggi}} = 1.0$, $\lambda_{\text{ocsf}} = 5.0$, and $\lambda_{\text{dptf}} = 1.0$.

\section{Details of Network Architecture}
\label{sec:supp_arch}
\subsection{Sparse BEV Fusion}
This section provides additional details on the sparse BEV fusion module (Section~3.2 of the main paper).
We construct the fused representation over the union of non-empty BEV cells from $F_G$ and $F_R$. 
At each cell $\mathbf{p}$, if one branch is inactive, its feature is set to a zero vector.
A lightweight gating network predicts channel-wise weights for the Gaussian and radar features.
The gates are defined as
\begin{equation}
    w_G(\mathbf{p}) = \sigma\!\left(\psi_G([F_G(\mathbf{p}), F_R(\mathbf{p})])\right), \quad
    w_R(\mathbf{p}) = \sigma\!\left(\psi_R([F_G(\mathbf{p}), F_R(\mathbf{p})])\right),
\end{equation}
where $\sigma(\cdot)$ is the sigmoid function, $[\cdot, \cdot]$ denotes channel-wise concatenation, and $\psi_G(\cdot)$, $\psi_R(\cdot)$ are Submanifold Sparse Convolution layers that produce channel-wise attention weights for the Gaussian and radar features, respectively.

The re-weighted features are concatenated as
\begin{equation}
    H(\mathbf{p}) = \bigl[w_G(\mathbf{p}) \odot F_G(\mathbf{p}),\ w_R(\mathbf{p}) \odot F_R(\mathbf{p})\bigr],
\end{equation}
where $\odot$ denotes element-wise multiplication. The concatenated feature $H$ is then processed by a sparse encoder $\mathcal{E}(\cdot)$, implemented as a residual block of $3 \times 3$ Submanifold Sparse Convolutions, to produce the fused sparse BEV feature map $F_{\mathrm{fuse}} = \mathcal{E}(H)$. The resulting $F_{\mathrm{fuse}}$ 
is passed to the subsequent DPTF module.

\subsection{Temporal BEV Aggregation}
\begin{center}
    \centering
    \includegraphics[width=0.99\textwidth]{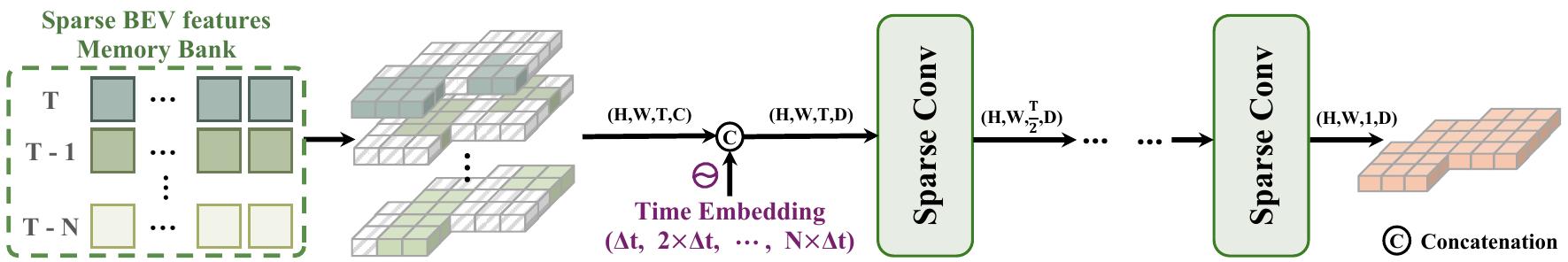}
    \captionof{figure}{Detailed architecture of the BEV-path temporal aggregation module.}
    \label{fig:bev_agg}
\end{center}
This section describes the BEV-path temporal aggregation module introduced in Section~3.3 of the main paper. 
As illustrated in Fig.~\ref{fig:bev_agg}, the temporally aligned sparse BEV features from the memory bank are stacked into a 4D sparse tensor. 
For each memory slot, the elapsed time $\Delta t_k$ is mapped to a time embedding via a lightweight MLP and concatenated with the corresponding features.
The resulting tensor is processed by a series of $3 \times 3 \times 3$ sparse convolutions that progressively reduce the temporal dimension, producing the temporally fused sparse BEV representation for the detection head.

\section{Implementation Details}
\label{sec:supp_impl}
\noindent{\textbf{Hyperparameters.}} We set the number of radar, camera, and random keypoints in KGGI to $K_r = 1024$, $K_c = 256$, and $K_{\text{rand}} = 256$, respectively. 
For the Gaussian path, we select 256 Gaussians per frame via farthest point sampling and maintain the past 4 frames, yielding $K_{\text{hist}} = 1024$ history Gaussians in total. 
The OCSF module uses $L = 2$ refinement layers, each with $K = 11$ sampling points, including 5 fixed and 6 learnable points, for deformable cross-attention. 
The splatting occupancy threshold is set to $\tau = 0.1$, the bounding box enlargement margin to $\epsilon = 4.8$\,m, and the BEV memory filtering threshold to $\tau_{\text{mem}} = 0.1$.

\noindent{\textbf{Data augmentation.}} For Image-view-space Data Augmentation (IDA), we apply horizontal flipping, scaling ($-0.06$ to $0.11$), and rotation ($\pm 5.4^\circ$). BEV-space Data Augmentation (BDA) includes random flipping along the $X$ and $Y$ axes, scaling ($0.95$--$1.05$), and rotation ($\pm 0.3925$ rad). We do not use ground-truth sampling augmentation.

\noindent{\textbf{Training configuration.}} Horizon3D is trained end-to-end in two stages. The single-frame model, consisting of KGGI and OCSF, is first trained for 24 epochs, followed by the full model including DPTF for an additional 24 epochs. 
We provide configurations for two image backbones: ResNet-50 with an input resolution of $256 \times 864$ and VoVNet-99 with $640 \times 960$.
For VoVNet-99, we apply a learning rate multiplier of $0.1$ to the image backbone.
Both configurations use AdamW with a base learning rate of $1 \times 10^{-4}$ and a cyclic learning rate schedule.
The detailed training hyperparameters are summarized in Table~\ref{tab:training_settings}.
\begin{table}[h]
\centering
\caption{Training settings for different backbone networks in Horizon3D.}
\label{tab:training_settings}
\begin{tabular}{l||c|c}
\toprule
\textbf{Configs} & \textbf{ResNet-50} & \textbf{VoVNet-99} \\
\midrule
Image Size & $256 \times 864$ & $640 \times 960$ \\
BEV Grid Size & {$384 \times 384$} & {$384 \times 384$} \\
Optimizer & AdamW & AdamW \\
Base Learning Rate & 1e-4 & 1e-4 \\
Weight Decay & 1e-2 & 1e-2 \\
Backbone LR Multiplier & 1.0 & 0.1 \\
Optimizer Momentum & $\beta_1, \beta_2=0.85, 0.95$  & $\beta_1, \beta_2=0.85, 0.95$  \\
Batch Size (Total) & 8 & 8 \\
Training Epochs (Phase 1 / Phase 2) & 24 / 24 & 24 / 24 \\
Temporal Frames (BEV / Gaussian) & 8 / 4 & 8 / 4 \\
LR Schedule & Cyclic & Cyclic \\
Gradient Clip & 35 & 35 \\
Hardware & 4$\times$ RTX 3090 & 4$\times$ RTX 5000 \\
\bottomrule
\end{tabular}
\end{table}
\begin{table}[t!]
\centering
\caption{Per-class AP comparison on the TruckScenes validation set (V2-99 backbone). 
'O.V.', 'Ped.', 'M.C.', 'T.C.', and 'T.S.' denote other vehicle, pedestrian, 
motorcycle, traffic cone, and traffic sign, respectively.}
\label{tab:per_class}
\resizebox{\textwidth}{!}{
    \begin{tabular}{l|c||cccccccccccc|c}
    \toprule[1.2pt]
    \textbf{Methods} & \textbf{Input} & \textbf{Car} & \textbf{Truck} & \textbf{Bus} & \textbf{Trailer} & \textbf{O.V.} & \textbf{Ped.} & \textbf{M.C.} & \textbf{Bicycle} & \textbf{T.C.} & \textbf{Barrier} & \textbf{Animal} & \textbf{T.S.} & \textbf{mAP} \\
    \midrule[0.4pt]
    Far3D \cite{Far3D} & C & 16.6 & 12.1 & 0.0 & 29.0 & 2.2 & 9.4 & 4.7 & 8.9 & 20.3 & 4.1 & 0.0 & 21.3 & 10.7 \\
    SpaRC \cite{SpaRC} & C+R & 43.8 & 29.6 & 1.1 & 40.0 & 8.6 & 18.3 & 16.5 & 14.1 & 40.8 & 14.3 & 0.0 & 41.9 & 22.5 \\
    \rowcolor[gray]{0.90}
    \textbf{Horizon3D (Ours)} & C+R & \textbf{44.3} & \textbf{32.9} & \textbf{6.9} & \textbf{41.1} & \textbf{2.7} & \textbf{28.5} & \textbf{25.7} & \textbf{30.6} & \textbf{32.6} & \textbf{9.1} & \textbf{0.0} & \textbf{33.7} & \textbf{24.1} \\
    \bottomrule[1.2pt]
    \end{tabular}
}
\end{table}
\FloatBarrier

\section{Additional Experimental Results}
\label{sec:supp_exp}
\subsection{Per-Class Performance on TruckScenes}
Table~\ref{tab:per_class} presents the per-class AP on the TruckScenes validation set. 
Horizon3D achieves the highest mAP of $24.1$, outperforming Far3D~\cite{Far3D} and SpaRC~\cite{SpaRC}, which obtain mAPs of $10.7$ and $22.5$, respectively.
Compared with SpaRC, Horizon3D shows notable AP gains on dynamic categories, including Pedestrian ($+10.2$), Motorcycle ($+9.2$), Bicycle ($+16.5$), Bus ($+5.8$), and Truck ($+3.3$), where our dual-path temporal fusion with velocity compensation is most effective.
However, Horizon3D shows lower AP on static categories, including Traffic Cone ($32.6$ vs.\ $40.8$), Barrier ($9.1$ vs.\ $14.3$), and Traffic Sign ($33.7$ vs.\ $41.9$).
All methods report $0.0$ AP on Animal due to the extreme scarcity of annotations in TruckScenes.

\subsection{Additional Ablation Studies}
Table~\ref{tab:ablation_epsilon} examines the effect of the enlargement margin $\epsilon$ used for occupancy supervision in OCSF (Section~3.2 of the main paper).
Performance improves as $\epsilon$ increases from $0.8$\,m to $4.8$\,m, peaking at $20.3$ mAP and $35.4$ NDS, and then decreases at $6.4$\,m.
A small margin restricts Gaussians to object interiors, limiting contextual encoding, while an excessively large margin introduces background noise.
We use $\epsilon = 4.8$\,m in all experiments.
\begin{table}[h]
\centering
\caption{Ablation study on the bounding box enlargement margin $\epsilon$.}
\label{tab:ablation_epsilon}
\begin{tabular}{c|c|cc}
\toprule
$\epsilon$ (m) & \# cells & mAP$\uparrow$ & NDS$\uparrow$ \\
\midrule
0.8 & 1 & 18.6 & 33.5 \\
1.6 & 2 & 19.8 & 34.1 \\
3.2 & 4 & 19.9 & 34.8 \\
\rowcolor[gray]{0.90}
\textbf{4.8} & \textbf{6} & \textbf{20.3} & \textbf{35.4} \\
6.4 & 8 & 19.6 & 34.7 \\
\bottomrule
\end{tabular}
\end{table}

\noindent\textbf{Effect of temporal self-attention.}
Table~\ref{tab:ablation_temp_attn} evaluates the effect of temporal self-attention within the Multimodal Gaussian Encoder (Section~3.3 of the main paper).
Adding temporal self-attention improves mAP and NDS by $+0.6$ and $+1.1$, respectively, as it enables cross-frame interaction among merged Gaussians beyond the local receptive field of sparse convolution and helps resolve residual misalignment after velocity compensation.
\begin{table}[h]
\centering
\caption{Ablation study on temporal self-attention in the Multimodal Gaussian encoder.}
\label{tab:ablation_temp_attn}
\begin{tabular}{c|cc}
\toprule
Temporal Self-Attn. & mAP$\uparrow$ & NDS$\uparrow$ \\
\midrule
 & 23.0 & 36.3 \\
\rowcolor[gray]{0.90}
\checkmark & \textbf{23.6} & \textbf{37.4} \\
\bottomrule
\end{tabular}
\end{table}

\noindent\textbf{Effect of the number of Multimodal Gaussian encoder layers.}
Table~\ref{tab:ablation_layers} studies the effect of the number of Multimodal Gaussian Encoder layers $L$ in the OCSF module. 
Increasing $L$ from $1$ to $2$ improves mAP and NDS by $+1.2$ and $+2.1$, respectively, while further increasing to $L = 4$ brings only marginal gains ($+0.7$ mAP) at more than twice the latency.
Since KGGI initializes Gaussians at estimated object keypoints, the primitives already start near foreground regions, largely bridging the gap between shallow and deeper encoder configurations. 
We adopt $L = 2$ as the default. All results are reported using the single-frame model (Phase 1).
\begin{table}[h]
\centering
\caption{Ablation study on the number of Multimodal Gaussian encoder layers $L$.}
\label{tab:ablation_layers}
\begin{tabular}{c|cc|c}
\toprule
$L$ & mAP$\uparrow$ & NDS$\uparrow$ & Latency (ms) \\
\midrule
1 & 19.2 & 33.3 & 14 \\
\rowcolor[gray]{0.90}
\textbf{2} & \textbf{20.3} & \textbf{35.4} & \textbf{31} \\
4 & 21.0 & 35.8 & 64 \\
\bottomrule
\end{tabular}
\end{table}

\noindent\textbf{Latency breakdown.}
Table~\ref{tab:latency} compares the module-wise latency of Horizon3D with BEV-based methods on a single RTX 3090. 
Horizon3D achieves a total latency of $117.7$\,ms, which is $2.0\times$ faster than BEVFusion~\cite{BEVFusion} ($233.4$\,ms) and $2.9\times$ faster than CRT-Fusion~\cite{CRT-Fusion} ($338.8$\,ms). 
The radar encoder and detection head both benefit from fully sparse processing, reducing their latency from $75$\,ms and $79$\,ms to $14.3$\,ms and $9.3$\,ms, respectively.

CRT-Fusion is slower than BEVFusion in view transform ($37.7$ vs.\ $11.1$\,ms) due to an additional radar-camera cross-attention module applied during view transformation, and in temporal fusion ($138.4$ vs.\ $60.8$\,ms) due to velocity compensation on the dense grid. 
Although CRT-Fusion stores only foreground cells, it reconstructs a dense grid for temporal aggregation and applies dense convolutions. 
Our DPTF ($36.8$\,ms) maintains a fully sparse representation throughout, performing temporal aggregation via sparse convolutions without dense reconstruction.

\section{Additional Qualitative Results}
\label{sec:supp_qual}
\begin{table}[t!]
    \centering
    \caption{Module-wise latency comparison (ms).}
    \label{tab:latency}
    \begin{minipage}[t]{0.50\linewidth}
        \centering
        \begin{tabular}{l|cc}
        \toprule
        Module & CRT-Fusion & BEVFusion \\
        \midrule
        Image encoder     & 7.5  & 7.2  \\
        Radar encoder     & 75.9  & 75.4  \\
        View transform    & 37.7  & 11.1  \\
        Temporal fusion   & 138.4 & 60.8  \\
        Detection head    & 79.3  & 78.9  \\
        \midrule
        \textbf{Total}    & \textbf{338.8} & \textbf{233.4} \\
        \bottomrule
        \end{tabular}
    \end{minipage}
    \hfill
    \begin{minipage}[t]{0.45\linewidth}
        \centering
        \begin{tabular}{l|c}
        \toprule
        Module & Horizon3D \\
        \midrule
        Image encoder     & 7.5  \\
        Radar encoder     & 14.3 \\
        KGGI              & 10.3 \\
        OCSF              & 39.5 \\
        DPTF              & 36.8 \\
        Detection head    & 9.3  \\
        \midrule
        \textbf{Total}    & \textbf{117.7} \\
        \bottomrule
        \end{tabular}
    \end{minipage}
\end{table}


































\subsection{Qualitative results of Horizon3D}
Fig.~\ref{fig:Supple_det} presents qualitative comparisons between CRT-Fusion and Horizon3D on the TruckScenes validation set. 
The figure demonstrates that Horizon3D produces more accurate detections (orange dashed regions), particularly at long range beyond $100$\,m where radar returns are sparse and objects appear small in camera images. 

\begin{center}
    \centering
    \includegraphics[width=0.99\textwidth]{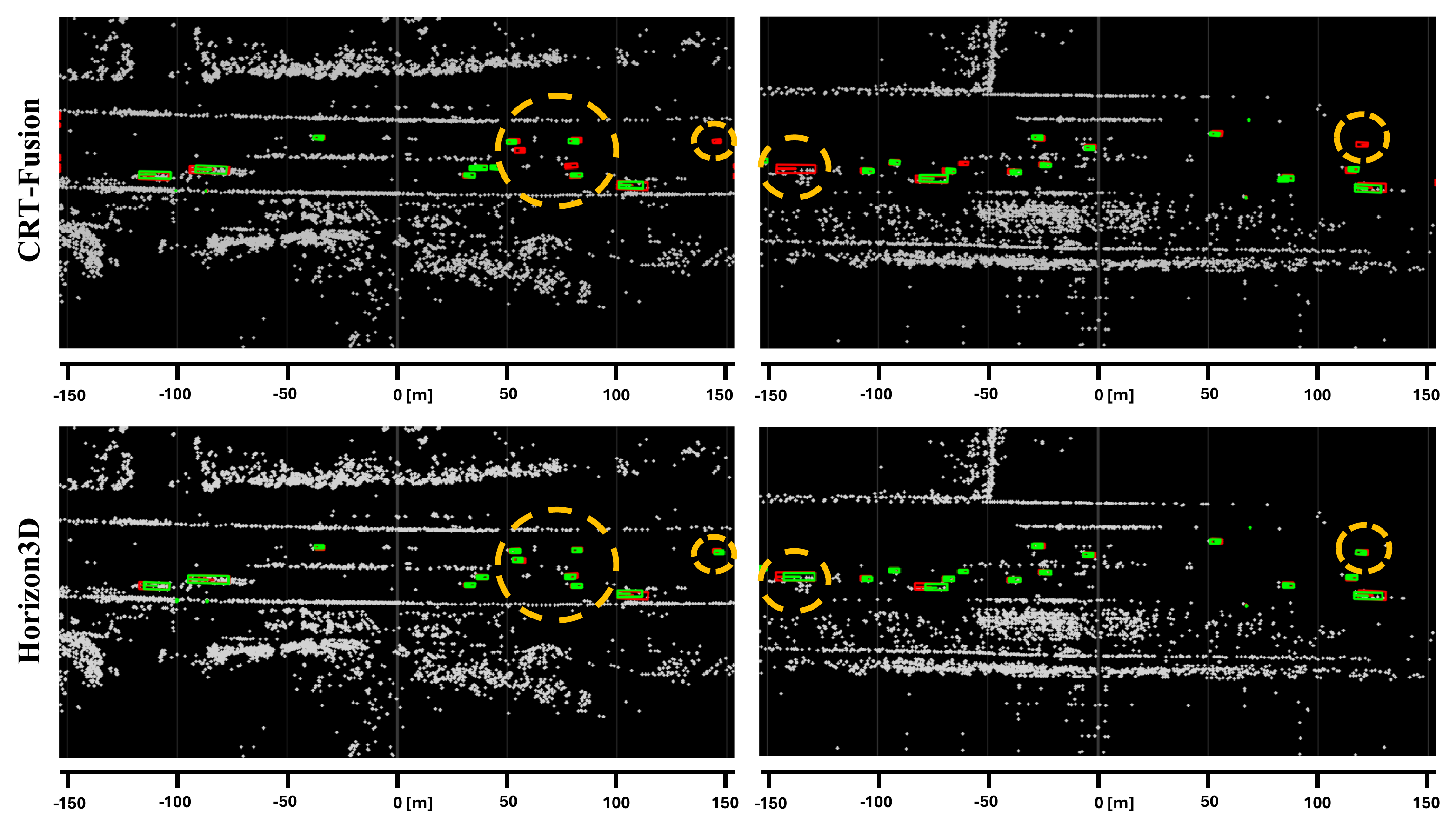}
    \captionof{figure}{\textbf{Qualitative comparison on TruckScenes validation set.} Green boxes are predictions and red boxes are ground truth. Gray dots represent radar point clouds. Orange dashed circles denote regions where Horizon3D produces more accurate detections.}
    \label{fig:Supple_det}
\end{center}

\subsection{Failure Cases}
Figure~\ref{fig:failure} illustrates failure cases of Horizon3D. 
On the left, the model struggles with other vehicles, whose elongated and continuous geometry is difficult to fully cover with a sparse set of Gaussians. 
On the right, missed detections are observed for traffic signs suspended above the roadway, where radar returns are rarely produced and Gaussians fail to form reliably. 
These observations are consistent with the per-class analysis in Table~\ref{tab:per_class}, where both categories show lower AP compared to SpaRC.

\begin{center}
    \centering
    \includegraphics[width=0.99\textwidth]{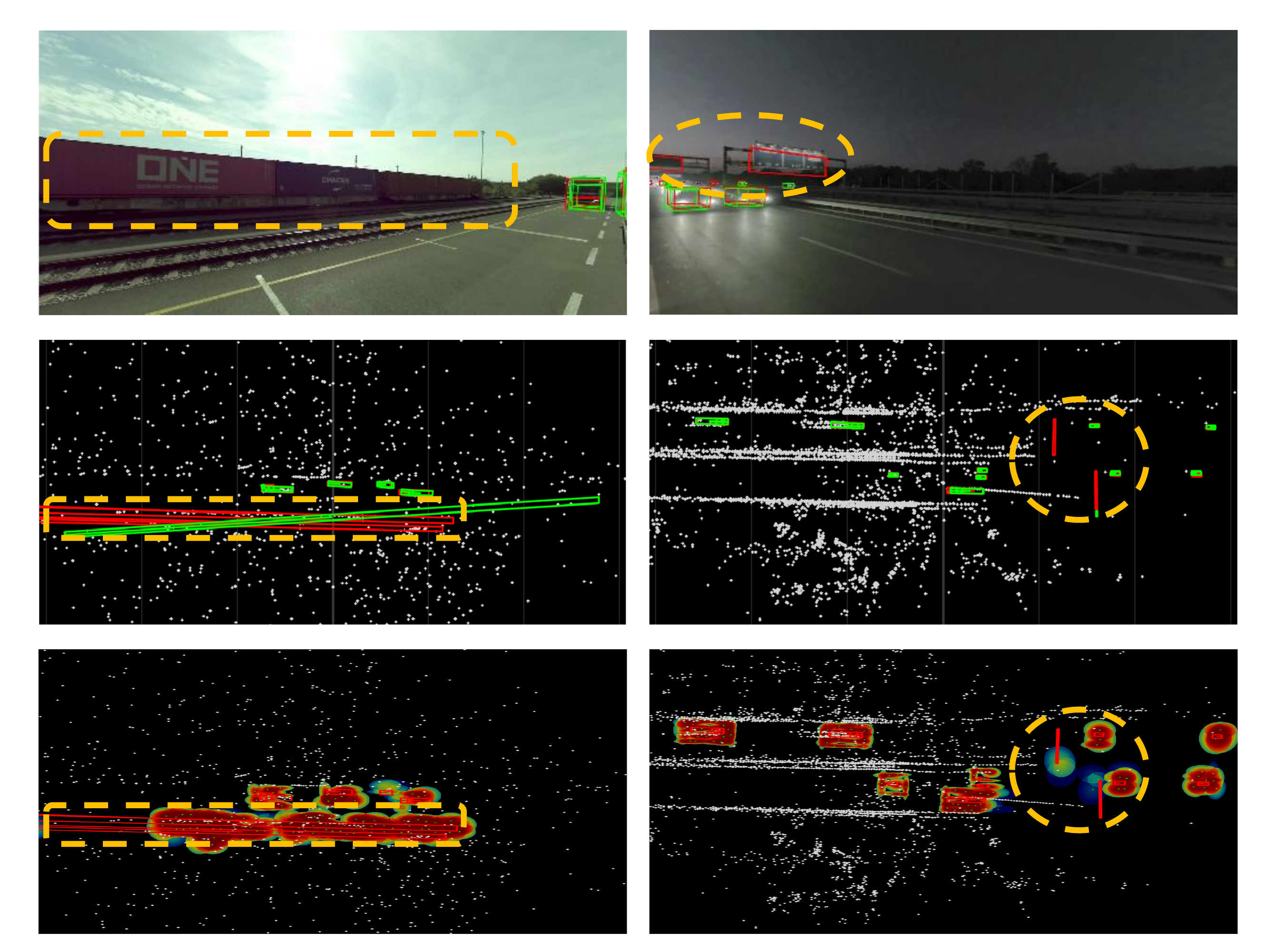}
    \captionof{figure}{\textbf{Failure cases on TruckScenes validation set.} The top row shows camera views, the middle row shows detection results on the BEV plane, and the bottom row visualizes each Gaussian primitive, where warmer colors indicate higher contributions during object-centric splatting and green dots denote Gaussian centers. Green boxes are predictions and red boxes are ground truth. Orange dashed circles highlight missed or inaccurate detections.}
    \label{fig:failure}
\end{center}

\endgroup
\end{document}